\pdfoutput=1
\documentclass[11pt,a4paper]{article}
\usepackage[hyperref]{naaclhlt2018}
\usepackage{times}
\usepackage{graphicx}
\usepackage{latexsym}
\usepackage{multirow}
\usepackage{url}
\usepackage{booktabs}
\aclfinalcopy 

\title{UBC-NLP at SemEval-2019 Task 6: \\ Ensemble Learning of Offensive Content With Enhanced Training Data}

\author{Arun Rajendran \qquad
  Chiyu Zhang \qquad
  Muhammad Abdul-Mageed \qquad
  \\Natural Language Processing Lab
  \\University of British Columbia
  \\\tt{{muhammad.mageed}@ubc.ca}}

\date{Affiliation}

\begin{document}
\pagestyle{plain}
\maketitle
\begin{abstract}
We examine learning offensive content on Twitter with limited, imbalanced data. For the purpose, we investigate the utility of using various data enhancement methods with a host of classical ensemble classifiers. Among the 75 participating teams in SemEval-2019 sub-task B, our system ranks 6th (with 0.706 macro F$_{1}$-score). For sub-task C, among the 65 participating teams, our system ranks 9th (with 0.587 macro F$_{1}$-score). 
\end{abstract}

\section{Introduction}
With the proliferation of social media, millions of people currently express their opinions freely online. Unfortunately, this is not without costs as some users fail to maintain the thin line between freedom of expression and hate speech, defamation, ad hominem attacks, etc. Manually detecting these types of negative content is not feasible, due to the sheer volume of online communication. In addition, individuals tasked with inspecting such types of content may suffer from depression and burnout. For these reasons, it is desirable to build machine learning systems that can flag offensive online content.

Several works have investigated detecting undesirable ~\cite{alshehri2018think} and offensive language online using traditional machine learning methods. For example, ~\citet{xiang2012detecting} employ statistical topic modelling and feature engineering to detect offensive tweets. Similarly, ~\citet{davidson2017automated} train multiple classifiers (e.g., logistic regression, decision trees, and support vector machines) to detect hate speech from general offensive tweets. More recently, deep artificial neural networks (i.e., deep learning) has been used for several text classification tasks, including detecting offensive and hateful language. For example, ~\citet{pitsilis2018detecting} use recurrent neural networks (RNN) to detect offensive language in tweets. ~\citet{mathur2018detecting} use transfer learning with convolutional neural networks (CNN) for offensive tweet classification on Twitter data.

Most of these works, however, either assume relatively balanced data (traditional classifiers) and/or large amounts of labeled data (deep learning). In scenarios where only highly imbalanced data are available, it becomes challenging to learn good generalizations. In these cases, it is useful to employ methods with good predictive power for especially minority classes. For example, methods capable of enhancing training data (e.g., by augmenting minority categories) are desirable in such scenarios. In the literature, some works have been undertaken to address issues of data imbalance in language tasks. For example, ~\citet{mountassir2012addressing} propose different undersampling techniques that yield better performance than common random undersampling on sentiment analysis. Along similar lines ~\citet{gopalakrishnan2014sentiment} propose a modified ensemble based bagging algorithm and sampling techniques that improve sentiment analysis. Further, ~\citet{li2018imbalanced} present a novel oversampling technique that generates synthetic texts from word spaces. 

In addition to data enhancement, combining various classifiers in an ensemble fashion can be useful since different classifiers have different learning biases. Past research has shown the effectiveness of ensembling classifiers for text classification ~\cite{xia2011ensemble,onan2016ensemble}. \linebreak ~\citet{omar2013ensemble}, for example, study the performance of ensemble models for sentiment analysis of Arabic reviews. ~\citet{da2014tweet} exploit ensembles to boost the accuracy on twitter sentiment analysis. ~\citet{wang2009diversity} demonstrate the utility of combining sampling techniques with ensemble models for solving the data imbalance problem. 

\begin{table*}[!htbp]
\centering
\begin{tabular}{@{}lll@{}}
\toprule
\textbf{Task}                       & \textbf{Label}      & \textbf{Example}                              \\ \midrule
\multirow{2}{*}{\textbf{Sub-task B}} & \textbf{targeted}   & \textit{Liberals are all Kookoo !!!}          \\
                                    & \textbf{untargeted} & \textit{Don’t believe the hype.}              \\ \midrule
\multirow{3}{*}{\textbf{Sub-task C}} & \textbf{individual} & \textit{Good move...he is the big loser}      \\
                                    & \textbf{group}      & \textit{The Liberals are mentally unstable!!} \\
                                    & \textbf{other}      & \textit{Google go to hell!}                   \\ \bottomrule
\end{tabular}
\caption{\label{tb:tweet-example} Examples of each class in sub-tasks B and C}
\end{table*} 

In this paper, we describe our submissions to SemEval-2019 task 6 (OffenseEval) \cite{offenseval}. We focus on sub-tasks B and C. The Offensive Language Identification Dataset \cite{OLID}, the data released by the organizers for each of these sub-tasks, is extremely imbalanced (see Section ~\ref{sec:data}). We propose effective methods for developing models exploiting the data. Our main contributions are: (1) we experiment with a number of simple data augmentation methods to alleviate class imbalance, and (2) we apply a number of classical machine learning methods in the context of ensembling to develop highly successful models for each of the competition sub-tasks. Our work shows the utility of the proposed methods for detecting offensive language in absence of budget for performing feature engineering and/or small, imbalanced data.

The rest of the paper is organized as follows: We describe the datasets in Section ~\ref{sec:data}. We introduce our methods in Section ~\ref{sec:methods}. Next, we detail our models for each sub-task (Sections ~\ref{sec:subtaskb} and ~\ref{sec:subtaskc}). We then offer an analysis of the performance of our models in Section ~\ref{sec:analysis}, and conclude in Section ~\ref{sec:conclusion}.

\section{Data}\label{sec:data}

As mentioned, \textit{OffenseEval} is SemEval-2019 \linebreak task 6. The task is focused on identifying and categorizing offensive language in social media and involves three different sub-tasks. These are:

\begin{itemize}
    \item \textbf{Sub-task A} is offensive language identification, e.g. classifying the given tweets into \textit{offensive} or \textit{non-offensive}. In our work, we only focus on sub-tasks B and C and so we do not cover sub-task A further. 
    \item \textbf{Sub-task B} is automatic categorization of offensive content types, which involves categorizing tweets into \textit{targeted} and \textit{untargeted} threats. The dataset for this sub-task consists of 4,400 tweets (3,876 \textit{targeted} and 524 \textit{untargeted}). Table ~\ref{tb:tweet-example} provides one examples of each of these two classes.

    \item \textbf{Sub-task C} is offense target identification and includes the 3 classes of targets. These classes are in the set \textit{\{individual, group, others\}}. The dataset for this sub-task consists of 3,876 tweets (2,407 \textit{individual}, 1,074 \textit{group}, and 395 \textit{other}). We similarly provide one example for each of these classes in Table ~\ref{tb:tweet-example}.
\end{itemize}

We use 80\% of the tweets as our training set and the remaining 20\% as our validation set for both sub-tasks B and C. We also report our best models on the competition test set, as returned to us by organizers. Table \ref{tb:data} provides statistics of our data for sub-tasks B and C. 

\begin{table}[!htbp]
\begin{tabular}{@{}llrrr@{}}
\toprule
\textbf{Task}                       & \textbf{Label}      & \multicolumn{1}{l}{\textbf{Train}} & \multicolumn{1}{l}{\textbf{Dev}} & \multicolumn{1}{l}{\textbf{Total}} \\ \midrule
\multirow{2}{*}{\textbf{Sub-task B}} & \textbf{targeted}   & 3,101                              & 775                              & 3,876                             \\
                                    & \textbf{untargeted} & 419                                & 105                              & 524                               \\ \midrule
\multirow{3}{*}{\textbf{Sub-task C}} & \textbf{individual} & 1,925                              & 482                              & 2,407                             \\
                                    & \textbf{group}      & 859                                & 215                              & 1,074                             \\
                                    & \textbf{other}      & 316                                & 79                               & 395                               \\ \bottomrule
\end{tabular}
\caption{\label{tb:data} Distribution of classes over our data splits}
\end{table}



\section{Methods}\label{sec:methods}
\subsection{Pre-Processing}
We utilize a simple data pre-processing pipeline involving lower-casing all text, filtering out URLs, usernames, punctuation, irrelevant characters and emojis, and splitting text into word-level tokens.

\subsection{Data Intelligence Methods}\label{subsec:sampling_tech}

We employ multiple machine learning methods and combine them with different sampling and data generation techniques to enhance our training set. From a data sampling perspective, the most common approaches to deal with imbalanced data is random oversampling and random undersampling ~\cite{lohr2009sampling,chawla2009data}. Learning with these basic techniques is usually effective due to possibly reducing model bias towards the majority class. We employ a number of data sampling techniques, as described next.

\textbf{Random oversampling} technique randomly duplicates the minority samples to obtain a balanced dataset. Despite the naive approach, this method is reported to perform well (as compared to other sophisticated oversampling methods) in the literature. One major drawback of this method is that it does not add any new data to the training set (since it only duplicates minority-class training data) \cite{liu2007generative}.

\textbf{Synthetic minority over-sampling (SMOTE)} is a sophisticated oversampling technique where synthetic samples are generated and added to the minority class. For each data point, one of \textit{k} minority class neighbours is randomly selected and the new synthetic point is a random point on the line joining the actual data point and this randomly selected neighbour. This method has been shown to be effective compared to some other oversampling methods ~\cite{chawla2002smote,Batista:2004:SBS:1007730.1007735}.

\textbf{Random undersampling} removes instances from the majority class in a random manner to obtain a balanced dataset. One possible disadvantage of this method is that it might remove valuable information from training data since, due to its randomness, it does not pay consideration to the data points removed ~\cite{liu2007generative}.

\textbf{kNN-based undersampling} is an alternative undersampling technique \cite{mani2003knn} which uses distance between points within a class. We use three different methods to select near-miss samples, as described in ~\citet{mani2003knn}. ~\textbf{NearMiss-1} selects majority class samples whose average distance to three closest minority class samples is smallest. In \textbf{NearMiss-2}, the samples of the majority class are selected such that their average distances to three farthest samples of minority class are smallest. \textbf{NearMiss-3} picks a given number of the closest majority class samples from each minority class sample, which guarantees every minority class sample is surrounded by some majority class points. ~\citet{mani2003knn} choose the majority class samples whose average distances to the three closest minority class samples are farthest. 

\textbf{Synthetic Data Generation.} We experiment with adding information to the minority class by generating synthetic samples employing a word2vec-aided paraphrasing technique. Initially, we train a word2vec model on the entire training data and use this word2vec model to generate samples for the minority class by randomly replacing words in tweets (with a probability of 0.9). We randomly pick one word from \textit{k} word2vec most similar words. We fix \textit{k=5} words and probability value as 0.9, but these are hyperparameters that can be optimized. 
In this way, we generate a balanced dataset in an attempt to overcome the problem of imbalance. In this technique, we draw inspiration from \cite{li2018imbalanced} where authors propose a sentiment lexicon generation method using a label propagation algorithm and utilize the generated lexicons to obtain synthetic samples for the minority class by randomly replacing a set of words with words that have similar semantic content.

\subsection{Classifiers}\label{subsec:classifier}
We apply a number of machine learning classifiers that are proven to work well for text categorization. Namely, we use logistic regression, support vector machines (SVM) and Naive Bayes. We also experiment with boosting algorithms such as random forest, AdaBoost, bagging classifier, \linebreak XGBoost, and gradient boosting classifier. We deploy ensembles of our best performing models in two ways: (1) ensembles based on majority rule classifiers that use predicted class labels for majority rule voting and (2) soft voting classifiers that predict the class label based on the argmax of the sums of the predicted probabilities of various classifiers.

\section{Sub-Task B Models}\label{sec:subtaskb}

For sub-task B, we have one minority class, so we generate samples for this minority class to obtain a new, balanced dataset. We use this balanced data as well as the the imbalanced (ORG) dataset for our first iteration of experiments. The goal of iteration is to identify the best (1) input n-gram settings (explained next), (2) classifier (from our classifiers listed in Section ~\ref{subsec:classifier}, and (3) sampling techniques (listed in ~\ref{subsec:sampling_tech}). For \textbf{n-gram settings}, we use a combination of bag of words and TF-IDF to extract features from the tweets and run with unigrams and all different combinations of unigram, bigrams, trigrams, and four grams. We run on all combinations across all the three variables above (n-grams, classifiers, and sampling methods) on both the imbalanced (ORG) and balanced datasets. Since our datasets are small, this iteration of experiments is not very costly. We acquire best results on the balanced dataset, identifying the combination of unigrams and bigrams as our best n-gram settings, XGBoost as the best classifier, and SMOTE as the best sampling technique. We provide these best results in Table ~\ref{tb:taskB-XgbResults} in Macro-F1 score. We use two baselines. Baseline 1 is the majority class in training data (i.e., \textit{targeted} offense class, 0.46827 Macro F$_{1}$-score). The second baseline is the best model with no data sampling, a logistic regression model. The best model, XGBoost with SMOTE sampling, acquires an F$_{1}$-score of 0.61248. This is a sizeable gain over the baselines. We now describe how we leverage ensembles to improve over this XGBoost model.

\begin{table}[!htbp]
\begin{tabular}{ccc}
\hline
\textbf{\begin{tabular}[c]{@{}c@{}}Sampling \\ Type\end{tabular}} & \textbf{\begin{tabular}[c]{@{}c@{}}Sampling \\ Technique\end{tabular}}   & \textbf{Macro F1} \\ \hline
\multirow{2}{*}{\textbf{NA}}                                                        & \textbf{Baseline 1}                                                        & 0.46827           \\ 
\cline{2-3}                                                        & \textbf{Baseline 2}                                                        & 0.5547           \\ \hline
\multirow{2}{*}{\textbf{Oversampling}}                            & \textbf{\begin{tabular}[c]{@{}c@{}}Random \\ Oversampling\end{tabular}}  & 0.56705           \\ \cline{2-3} 
                                                                  & \textbf{SMOTE}                                                           & \textbf{0.61248}           \\ \hline
\multirow{4}{*}{\textbf{Undersampling}}                           & \textbf{\begin{tabular}[c]{@{}c@{}}Random \\ Undersampling\end{tabular}} & 0.49739           \\ \cline{2-3} 
                                                                  & \textbf{Near Miss-1}                                                     & 0.32533           \\ \cline{2-3} 
                                                                  & \textbf{Near Miss-2}                                                     & 0.4376            \\ \cline{2-3} 
                                                                  & \textbf{Near Miss-3}                                                     & 0.46158           \\ \hline
\end{tabular}
\caption{\label{tb:taskB-XgbResults} \textbf{Sub-Task B:} XGBoost performance with sampling methods. Baseline 1 is our majority class in training data. Baseline 2 is a logistic regression model with no data sampling.}
\end{table}

\subsection{Ensembles for Sub-Task B}

\begin{table*}[!htbp]
\footnotesize
\centering
\begin{tabular}{ccccccccc}
\hline
\multirow{2}{*}{\textbf{Dataset}} & \multirow{2}{*}{\textbf{Models}} & \multicolumn{3}{c}{\textbf{Targeted}}              & \multicolumn{3}{c}{\textbf{Untargeted}}            & \textbf{}                                                          \\ \cline{3-9} 
                                  &                                  & \textbf{Precision} & \textbf{Recall} & \textbf{F1} & \textbf{Precision} & \textbf{Recall} & \textbf{F1} & \textbf{\begin{tabular}[c]{@{}l@{}}Macro \\ F1 score\end{tabular}} \\ \hline
\multirow{3}{*}{\textbf{DEV}}     & \textbf{XGBoost (SMOTE)}         & 0.90158            & \textbf{0.95742}         & \textbf{0.92866}     & \textbf{0.42105}            & 0.22857         & 0.2963      & 0.61248                                                            \\ \cline{2-9} 
                                  & \textbf{Model A}           & 0.90945            & 0.89419         & 0.90176     & 0.30508            & 0.34286         & \textbf{0.32287}     & 0.61231                                                            \\ \cline{2-9} 
                                  & \textbf{Model B}      & \textbf{0.94065}            & 0.90447         & 0.9222      & 0.26667            & \textbf{0.37838}         & 0.31285     & \textbf{0.61753}                                                            \\ \hline \hline
\multirow{3}{*}{\textbf{TEST}}    & \textbf{XGBoost (SMOTE)}          & 0.9004             & \textbf{0.9765}          & 0.9369      & 0.4444             & 0.1481          & 0.2222      & 0.57958                                                            \\ \cline{2-9} 
                                  & \textbf{Model A}           & \textbf{0.9378}             & 0.9202          & 0.9289      & 0.4516             & \textbf{0.5185}          & \textbf{0.4828}      & \textbf{0.70583}                                                            \\ \cline{2-9} 
                                  & \textbf{Model B}      & 0.9079             & 0.9718          & \textbf{0.9388}      & \textbf{0.5000}                & 0.2222          & 0.3077      & 0.62323                                                            \\ \hline
\end{tabular}
\caption{\label{tb:taskB-Results} \textbf{Sub-Task B:} Best ensemble model results. We reproduce XGBoost results from Table ~\ref{tb:taskB-XgbResults} for comparison.}
\end{table*}
Our best performance with the XGBoost model in the previous section was acquired with SMOTE oversampling. However, we note that oversampling in general performed better than other sampling methods. For this reason, we experiment with a number of ensemble methods across our two oversampling techniques (SMOTE and random oversampling [ROS]). We provide our best results from this iteration of experiments (for both the dev and the competition test set) in Table ~\ref{tb:taskB-Results}. In addition to the same XGBoost model reported earlier (in Table ~\ref{tb:taskB-XgbResults}, reproduced in Table ~\ref{tb:taskB-Results}), we identify and report our two best models: (1) \textbf{Model A}: An ensemble with soft voting over XGBoost, AdaBoost, and logistic regression with random oversampling (ROS) and (2) \textbf{Model B}: The average of our XGBoost model (with SMOTE) and the best model with synthetic oversampling (which is a Naive Bayes classifier). We submitted the three models in Table  ~\ref{tb:taskB-Results} to the competition. Although Model B performs best on the dev set, it was model A that performed highest on the competition test set. This suggests that the dev and test sets are different in some aspects. Importantly, even though the three models in Table ~\ref{tb:taskB-Results} perform comparably on dev, only the ensemble models (Model A and Model B) seem to generalize better on the test set. This further demonstrates the utility of ensembles on the task.

\section{Sub-Task C Models}\label{sec:subtaskc}
Sub-Task C is 3-way classification, with 2 minority classes. Again, we run all our classifiers with unigram and bigram combinations across all sampling methods (including no sampling) on this imbalanced dataset. In addition, we use 4 different configurations to generate samples for each of the two minority classes to obtain 4 balanced datasets. \texttt{C1} is created with random oversampling of the two minority classes; \texttt{C2} is created with synthetic oversampling of the two minority classes; \texttt{C3} is created with random oversampling of minority class \textit{group} (GRP) and synthetic oversampling of minority class \textit{other} (OTH); and \texttt{C4} is random oversampling of minority class OTH and synthetic oversampling of minority class GRP. 

We report our best results in Table ~\ref{tb:IndModel-taskC-Results}, with two baselines: Baseline 1 is the majority class in training data and Baseline 2 is our best model without sampling (a logistic regression classifier). Our best model on C2 is a logistic regression classifier, whereas our best models on C1, C3, and C4 are acquired with the same soft voting ensemble in Table ~\ref{tb:taskB-Results} (an ensemble of logistic regression, AdaBoost, and XGBoost). 


\begin{table}[]
\begin{tabular}{llll}
\hline
\multicolumn{1}{c}{\textbf{Sampling}}                     & \textbf{\begin{tabular}[c]{@{}l@{}}Type\end{tabular}} & \textbf{Best Model} & \textbf{Macro F1} \\ \hline
\multicolumn{1}{c}{\multirow{2}{*}{\textbf{NA}}} & NA                                                                    & Baseline-1          & 0.21300           \\ \cline{2-4} 
\multicolumn{1}{c}{}                                      & NA                                                                    & Baseline-2          & 0.51580           \\ \hline
\multirow{4}{*}{\textbf{Sampling}}                        & \textbf{C1}                                                           & \textbf{Model 1}    & \textbf{0.56822}  \\ \cline{2-4} 
                                                          & C2                                                                    & Log Reg             & 0.54319           \\ \cline{2-4} 
                                                          & C3                                                                    & Model 1             & 0.54665           \\ \cline{2-4} 
                                                          & C4                                                                    & Model 1             & 0.56216           \\ \hline
\end{tabular}
\caption{\label{tb:IndModel-taskC-Results} \textbf{Sub-Task C:} Best results with various sampling methods.} 
\end{table}


Our next step is to investigate whether we can further improve performance by averaging classification probabilities of models described in Table ~\ref{tb:IndModel-taskC-Results}. The result of this iteration is shown in Table ~\ref{tb:taskC-Results}. Models in Table ~\ref{tb:taskC-Results} are the 3 models we submitted to the SemEval-2019 competition, which are as follows: \textbf{Model 1}: our best model with C1; \textbf{Model 2}: a prediction based on the average of classification probabilities of the best classifiers on C1, C2, and C4; \textbf{Model 3}: the prediction acquired from the average of tag probabilities of the best classifiers on C1 and C4. Table \ref{tb:taskC-Results} shows that performance of all the models on the dev set is very comparable, with model 3 performing slightly better than the two other models. Similarly, results of the three models are not very different on the competition test set. 

\begin{table*}[!htbp]
\centering
\captionsetup{justification=centering}
\begin{tabular}{cccccc}
\hline
\textbf{Dataset}                                   & \textbf{Models} & \textbf{GRP}     & \textbf{IND}     & \textbf{OTH}     & \textbf{Macro F1 score} \\ \hline
\multirow{3}{*}{\textbf{DEV}}                      & Model 1         & 0.60538          & \textbf{0.82476} & 0.27451          & 0.56822                 \\ \cline{2-6} 
                                                   & Model 2         & \textbf{0.61504} & 0.8204           & 0.28931          & 0.57492                 \\ \cline{2-6} 
                                                   & Model 3         & 0.61207          & 0.8203           & \textbf{0.29577} & \textbf{0.57605}        \\ \hline \hline
\multicolumn{1}{l}{\multirow{3}{*}{\textbf{TEST}}} & Model 1         & \textbf{0.7101}  & 0.8116           & 0.2400             & \textbf{0.58722}        \\ \cline{2-6} 
\multicolumn{1}{l}{}                               & Model 2         & 0.686            & 0.8098           & 0.2041           & 0.56663                 \\ \cline{2-6} 
\multicolumn{1}{l}{}                               & Model 3         & 0.6946           & \textbf{0.819}   & \textbf{0.2449}  & 0.58619                 \\ \hline
\end{tabular}
\caption{\label{tb:taskC-Results} \textbf{Sub-Task C:} Results of our 3 final submitted models}
\end{table*}

\section{Model Analysis}\label{sec:analysis}

In order to further understand the results on the test set, we investigate the predictions made by our models across the two sub-tasks. For the purpose, we provide simple visualizations of the confusion matrices of predictions acquired by our best models as released by organizers.

\begin{figure}
  \centering
  \captionsetup{justification=centering}
  \includegraphics[width=\linewidth]{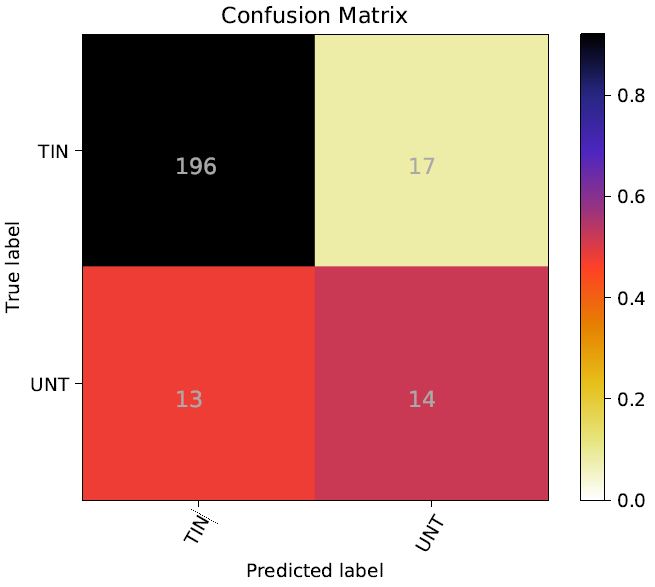}
  \caption{Confusion matrix of soft voting \\ensemble model (Model A in Table ~\ref{tb:taskB-Results}) for Sub-Task B.}
  \label{fig:conf_matrix_taskb}
\end{figure}

\textbf{Sub-Task B.} Figure ~\ref{fig:conf_matrix_taskb} shows that our model has higher precision for the targeted threats, which is also clear from Table ~\ref{tb:taskB-Results} presented earlier. Figure ~\ref{fig:conf_matrix_taskb} also shows that our model has slightly higher false negatives as compared to false positives. In other words, the chances of our model mislabeling a \textit{targeted} tweet as \textit{untargeted} is slightly higher as compared to predicting an \textit{untargeted} tweet as \textit{targeted}. 

\begin{figure}
  \centering
  \captionsetup{justification=centering}
  \includegraphics[width=\linewidth]{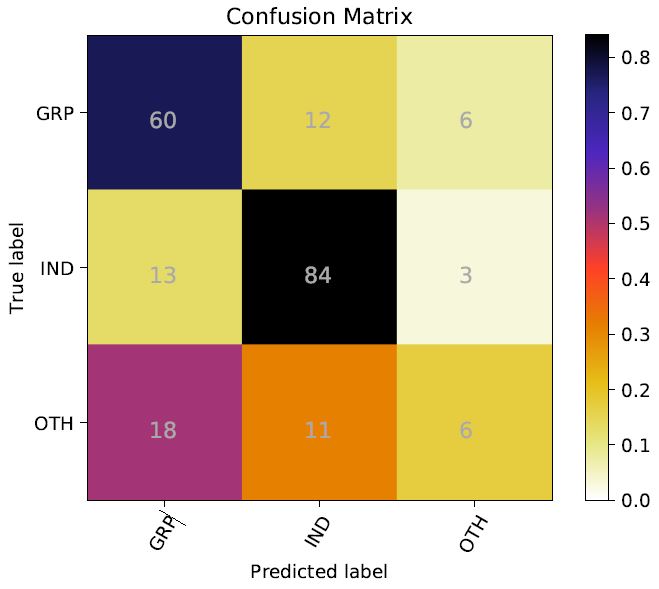}
  \caption{Confusion matrix of soft voting \\ensemble model (Model 1 in Table ~\ref{tb:taskC-Results}) for Sub-Task C.}
  \label{fig:conf_matrix_taskc}
\end{figure}

\textbf{Sub-Task C}
We visualize model errors in Figure ~\ref{fig:conf_matrix_taskc}. Figure ~\ref{fig:conf_matrix_taskc} shows that our model has higher precision for the \textit{group} (GRP) and \textit{individual} (IND) categories, but only higher recall for the \textit{other} (OTH) class. Again, this means that the chances of our model predicting a GRP tweet or IND tweet as OTH is much higher as compared to OTH tweet being predicted as IND or GRP. In other words, the model is biased towards predicting one of the two categories GRP and IND 
 
\section{Conclusion}\label{sec:conclusion}

In this paper, we described our contributions to OffenseEval, the 6th shared task of SemEval-2019 . We explored the effectiveness of different sampling techniques and ensembling methods combined with different classical and boosting machine learning algorithms. We find simple data enhancement approaches (i.e., sampling techniques) to work well, especially when coupled with the right ensemble methods. In general, ensemble models decrease errors by leveraging the different strengths of the various underlying models and hence are useful in absence of balanced data. 

\section{Acknowledgement}
We acknowledge the support of the Natural Sciences and Engineering Research Council of Canada (NSERC) and the Social Sciences Research Council of Canada (SSHRC). The research was partially enabled by WestGrid (\url{www.westgrid.ca}) and Compute Canada (\url{www.computecanada.ca}).


\bibliography{semeval2018}
\bibliographystyle{acl_natbib}

\end{document}